\documentclass[conference]{IEEEtran}
\IEEEoverridecommandlockouts
\usepackage{cite}
\usepackage{amsmath,amssymb,amsfonts}
\usepackage{algorithmic}
\usepackage{graphicx}
\usepackage{textcomp}
\usepackage{xcolor}
\usepackage{balance}

\def\BibTeX{{\rm B\kern-.05em{\sc i\kern-.025em b}\kern-.08em
    T\kern-.1667em\lower.7ex\hbox{E}\kern-.125emX}}
\begin{document}

\title{Unconscious and Intentional Human Motion Cues for Expressive Robot-Arm Motion Design}


\author{\IEEEauthorblockN{Taito Tashiro}
\IEEEauthorblockA{\textit{School of Engineering} \\
\textit{University of Hyogo}\\
Himeji, Hyogo, Japan \\
eo21f066@guh.u-hyogo.ac.jp}
\and
\IEEEauthorblockN{Tomoko Yonezawa}
\IEEEauthorblockA{\textit{Faculty of Informatics} \\
\textit{Kansai University}\\
Takatsuki, Osaka, Japan \\
yone@kansai-u.ac.jp}
\and
\IEEEauthorblockN{Hirotake Yamazoe}
\IEEEauthorblockA{\textit{Graduate School of Engineering} \\
\textit{University of Hyogo}\\
Himeji, Hyogo, Japan \\
yamazoe@eng.u-hyogo.ac.jp}
\and
}

\maketitle

\begin{abstract}
This study investigates how human motion cues can be used to design expressive robot-arm movements. Using the imperfect-information game \emph{Geister}, we analyzed two types of human piece-moving motions: natural gameplay (unconscious tendencies) and instructed expressions (intentional cues). Based on these findings, we created phase-specific robot motions by varying movement speed and stop duration, and evaluated observer impressions under two presentation modalities: a physical robot and a recorded video. Results indicate that late-phase motion timing, particularly during withdrawal, plays an important role in impression formation and that physical embodiment enhances the interpretability of motion cues. These findings provide insights for designing expressive robot motions based on human timing behavior.
\end{abstract}


\section{Introduction}
Robots are increasingly integrated into daily life, operating in schools, hospitals, homes, and public spaces. While social robots commonly use explicit expressive channels such as facial expressions or gestures, task-oriented robots—particularly robotic arms—rarely incorporate such mechanisms for conveying intention. Prior studies have shown that a lack of expressive behavior can hinder user acceptance and smooth collaboration~\cite{deGraaf2017,Fink2013}, and that enabling a robot to make its intention legible can improve coordination and trust in human–robot interaction~\cite{breazeal2005}.

A promising approach for enabling robots to convey intention is through motion imitation. Designing robot movements based on human motion cues can support intuitive, non-verbal communication~\cite{Fu2024,Gao2024,Nguyen2021}. Board games, with their well-defined rules and constrained action spaces, have been widely used in HRI research as reproducible settings for analyzing non-verbal behavior and decision-making processes~\cite{ray2022,zhang2024}.

We use the imperfect-information game \emph{Geister} as our experimental setting because it requires players to act under hidden strategic intention. Each piece has a strategic role: red pieces (evil ghosts) are intentionally offered to the opponent, whereas blue pieces (good ghosts) must be protected. Thus, even when performing the same action of moving a piece, players exhibit subtle differences in motion that reflect their intention. This makes \emph{Geister} suitable for analyzing how intention is communicated through motion cues. Our prior work in this setting examined how adding a pause influenced impressions of robot motion~\cite{koga2024}, but it remained unclear how human timing behaviors—specifically movement speed and stop duration—influence such impressions and can serve as intention cues for expressive robot-arm motion design.

\begin{figure}[tb]
    \centering
    \includegraphics[width=4.5cm]{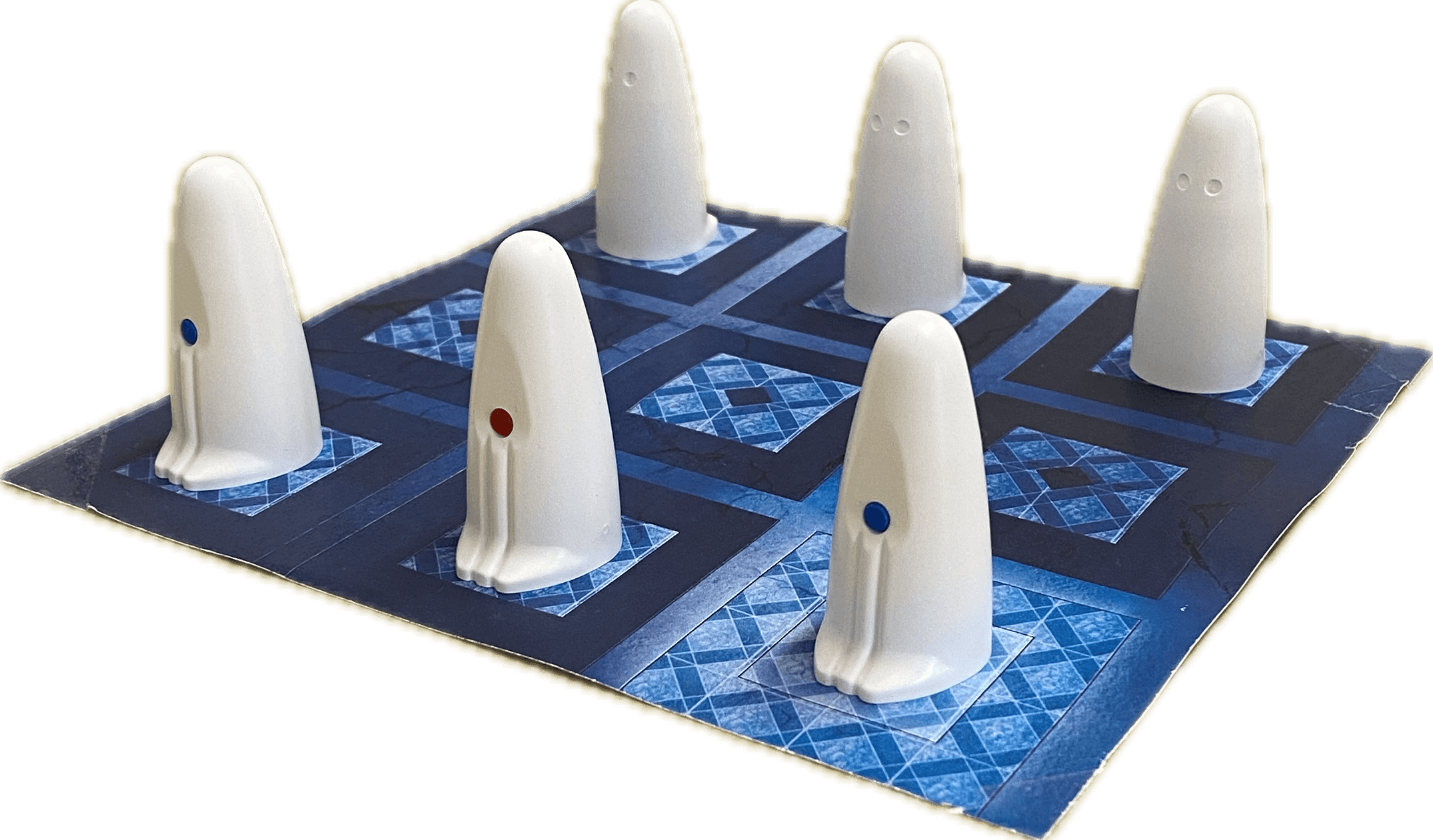}
    \caption{Pieces and Board of the Simplified Geister Game}
    \label{fig:board}
\end{figure}

In this study, we analyze human arm movements in simplified \emph{Geister} (Fig.~\ref{fig:board}) to investigate how timing parameters—movement speed and stop duration—affect the expression of intention in piece manipulation. We compare two types of human motion: unconscious movements that naturally occur during gameplay and intentionally emphasized movements produced to communicate intention. Based on these findings, we design expressive robot-arm motions by manipulating timing parameters and evaluate their perceptual effects.

A subset of our findings was previously reported~\cite{HAI2025poster}, including the analysis of unconscious human movements (Exp.~1-1) and the evaluation of robot motions presented physically (Exp.~2-1). This paper extends that work in two ways: (1) by analyzing intentionally produced movements to identify intention-related motion cues (Exp.~1-2), and (2) by examining how presentation modality—physical versus video presentation—affects impressions of robot motion (Exp.~2-2).

Although this study uses a game setting, conveying intention through timing-based motion design is broadly applicable to collaborative HRI, where trust and task efficiency are essential~\cite{webb2024co}.

\section{Simplified Geister}
We used a simplified version of \emph{Geister} to shorten match duration and facilitate efficient motion data collection for piece manipulation analysis. While the original game is played on a $6{\times}6$ board with eight pieces per player, we adopted the $3{\times}3$ version with two blue (good) pieces and one red (evil) piece per player, as shown in Fig.~\ref{fig:board}. Prior work has reported that this simplification retains the core strategic structure of the original game~\cite{matsuzaki2021}.
The basic rules are preserved and are summarized as follows:
\begin{itemize}
  \item Win by capturing both opponent blue pieces
  \item Lose by capturing the opponent red piece
  \item Win by escaping a blue piece through an escape square
\end{itemize}

\section{Experiment 1: Human Movement Analysis}
Experiment~1 investigates how human arm movements reflect differences in intention during piece manipulation. It consists of two parts: Exp.~1-1 analyzes unconscious movements that naturally occur during gameplay, and Exp.~1-2 analyzes intentionally emphasized movements produced to convey intention. By comparing these two types of human motion, we examine how intention influences timing-based motion cues.

\subsection{Experimental Setup}
\begin{figure}[tb]
    \centering
    \includegraphics[width=4.5cm]{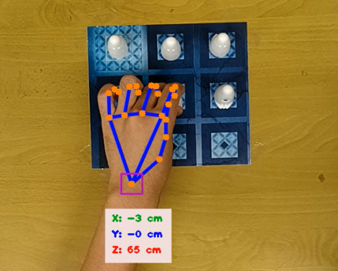}
    \caption{Example of Data Acquisition}
    \label{fig:handcapture}
\end{figure}

Participants’ arm movements were captured using an RGB-D camera (Luxonis OAK-D S2) connected to a desktop PC, as shown in Fig.~\ref{fig:handcapture}. Although the system can also estimate finger movements, we extracted only the wrist trajectory to represent arm movement.

The same recording setup was used in both Exp.~1-1 (unconscious movements during gameplay) and Exp.~1-2 (intentionally emphasized movements). The recorded wrist trajectories were segmented into six motion phases—approach, grasp, lift, placement, pre-release stop, and withdrawal. For each phase, timing-based motion cues, such as movement speed and stop duration, were computed for subsequent analysis.

\subsection{Exp.~1-1: Natural Piece-Moving Motions}
The purpose of Exp.~1-1 was to collect natural piece-moving motions that emerge during gameplay. Participants played the simplified version of \emph{Geister} in pairs while their wrist trajectories were recorded. Four male participants in their 20s took part in the experiment, all of whom had prior experience with the game. Each pair played multiple matches, resulting in a total of 10 recorded games. The average duration of a match was 135 seconds. No instructions regarding arm movements were given to preserve natural gameplay behavior.

\begin{figure}[tb]
    \centering 
    \includegraphics[width=0.9\hsize]{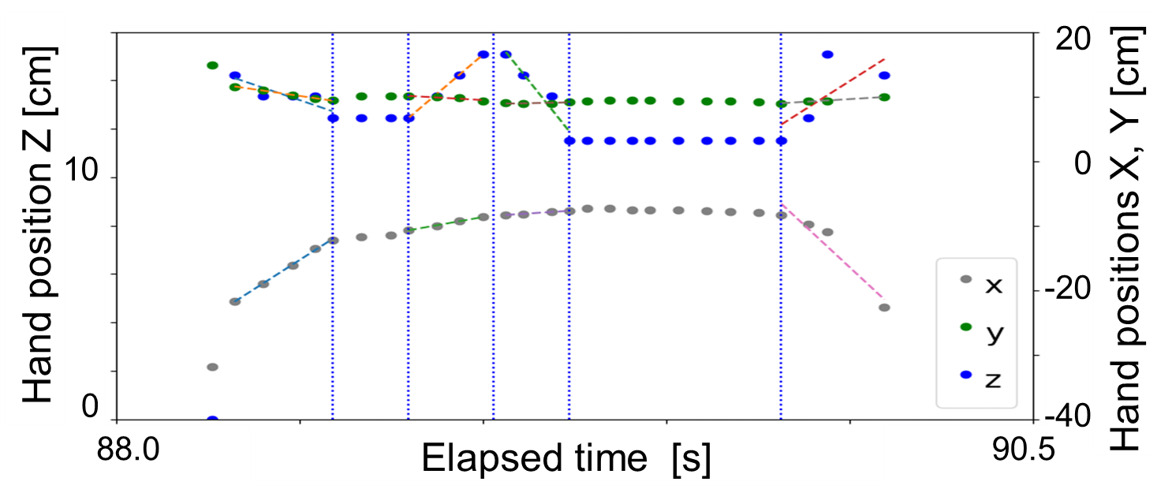}
    \label{fig:handcapture_results}
    \vspace{-5mm}
    \caption{Representative wrist trajectory during game-piece manipulation. Elapsed time (s) on the horizontal axis; 3D hand position on the vertical plots with $x$ = front–back, $y$ = left–right, and $z$ = height from the board. Blue dashed lines indicate the six phases (Phase 1–6).}
    \label{fig:trajectory}
\end{figure}

\noindent\textbf{Results:}
Figure~\ref{fig:trajectory} shows a representative wrist trajectory during piece manipulation. A total of 29 forward moves (18 blue, 11 red) from valid trials were analyzed. Each trajectory was segmented into six motion phases (Phase 1–6; see Fig.~\ref{fig:trajectory}). For each phase, movement speed and stop duration were extracted, and paired $t$-tests were conducted. Blue-piece moves tended to be slower during placement (Phase 4, $p<.1$), showed shorter pre-release stops (Phase 5, $p<.05$), and exhibited faster withdrawal (Phase 6, $p<.1$). These results suggest that timing-based intention cues are concentrated in the later motion phases.

\subsection{Exp.~1-2: Intentionally Expressed Piece-Moving Motions}
Exp.~1-2 was conducted to collect intentionally emphasized piece-moving motions. Eight male participants in their 20s were instructed to move pieces forward while explicitly expressing the following intentions:

\begin{itemize}
\setlength{\leftskip}{-2mm}
\item Red: convey ``wanting the opponent to capture the piece''
\item Blue: convey ``not wanting the opponent to capture the piece''
\end{itemize}

Participants first rehearsed the motions mentally and were allowed to plan freely how to express the instructed intentions. They then performed one motion for each condition (red and blue), and wrist trajectories were recorded. Oral informed consent was obtained from all participants.

\noindent\textbf{Results:}
Since the task involved subjective expression, large individual differences were observed. Therefore, statistical testing was not applied; instead, we analyzed consistent trends by counting participants whose motion parameters (speed or stop duration) differed by a factor of two or more between red and blue pieces. Several phase-specific tendencies were observed across participants—for example, faster approach in Phase 1 for blue pieces and longer stop duration in Phase 2 for red pieces (Table~\ref{tab:result2}). These results indicate that when explicitly instructed to express intention, participants tended to exaggerate timing in specific motion phases, producing clearer cues than in natural gameplay.

\begin{table}[tb]
    \caption{Observed Trends in Exp.~1-2}
    \centering
    \label{tab:result2}
    \begin{tabular}{cll}
        \hline
        Phase & & Observed Trend\\
        \hline
        \hline
        1 & Speed & Blue $>$ Red (majority)\\
        2 & Duration & Blue $<$ Red (majority)\\
        3 & Speed & Blue $>$ Red / Blue $<$ Red (split)\\
        4 & Speed & Blue $>$ Red / Blue $<$ Red (split)\\
        5 & Duration & Blue $>$ Red / Blue $<$ Red (split)\\
        6 & Speed & Blue $<$ Red (majority)\\
        \hline
    \end{tabular}
\end{table}

\subsection{Discussion of Experiment 1}
Comparing Exp.~1-1 and Exp.~1-2 highlights a key difference between unconscious and intentional expression in human movement. In natural gameplay (Exp.~1-1), players often attempted to conceal their intentions for strategic reasons, which reduced observable differences between red and blue piece movements. Even so, subtle timing tendencies emerged in the later motion phases, indicating implicit hesitation or confidence.

In contrast, intentional expression (Exp.~1-2) produced clearer and phase-specific differences, although with large individual variability. These findings suggest that unconscious human motion patterns provide weak but consistent timing cues, whereas intentionally emphasized movements yield stronger and more interpretable signals.

This distinction informs the motion design in Experiment 2, where both types of human-derived cues are used to generate robot-arm motions and evaluate how effectively they convey intention to observers.

\section{Experiment 2: Robot Motion Evaluation}
\subsection{System Configuration}
The robot arm used in this experiment was the myCobot 280M5 (Elephant Robotics), equipped with a dedicated gripper for piece manipulation. A desktop PC controlled both the robot arm and an RGB-D camera (Luxonis OAK-D S2), which was used to estimate the target piece's position before each motion.

We implemented seven robot-arm motion patterns: one baseline condition (slow movements with long stops) and six comparative conditions, each modifying a single phase of the motion by either increasing speed or shortening stop duration. The baseline motion and parameter variations were designed based on the human movements captured in Experiment 1-1. In all cases, the motion involved advancing a piece one square forward from the center of the board’s front row. The robot motions were generated using MoveIt in ROS, with the velocity parameter set to 35 (slow) or 50 (fast), and the stop duration to 2.0 s (long stop) or 0.2 s (short stop).

\begin{figure}[tb]
\centering
\begin{minipage}{0.45\linewidth}  
    \centering
    \includegraphics[width=0.8\linewidth]{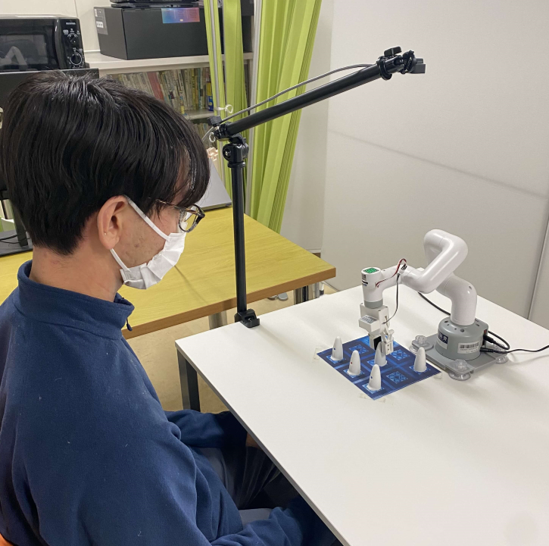}\\
    {\footnotesize (a) Robot Condition}
    \label{fig:env1}
\end{minipage}
\begin{minipage}{0.45\linewidth}  
    \centering
    \includegraphics[width=0.8\linewidth]{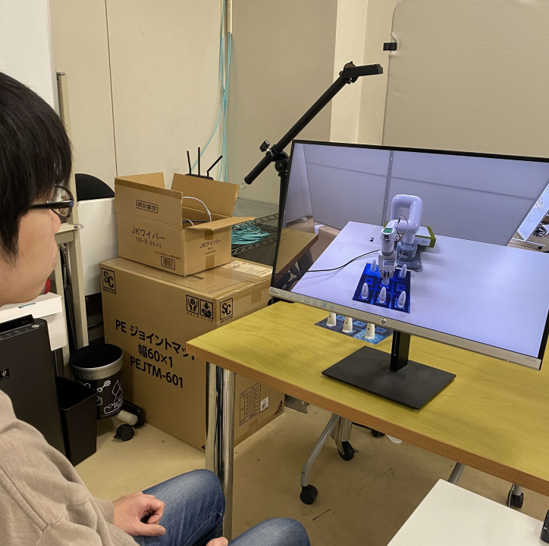}\\
    {\footnotesize (b) Video Condition}
    \label{fig:env2}
\end{minipage}
\caption{Experimental Environments for Each Presentation Condition}  
\label{fig:env}
\end{figure}

We manipulated only timing parameters while keeping the spatial trajectory constant to isolate the perceptual effects of temporal motion cues and examine how timing variations influence perceived intention. The experimental environments for both the robot presentation condition and video presentation condition are illustrated in Fig.~\ref{fig:env}.

\subsection{Exp.~2-1: Robot Condition}
The purpose of Exp.~2-1 was to evaluate how robot motion timing influences observer impressions when the motions are presented in physical form. Participants directly observed the robot performing the seven motion patterns. Ten participants (five male, five female; aged 20–30) took part in the experiment.

Each participant completed 12 trials, each consisting of a pairwise comparison between the baseline motion and one comparative motion. The presentation order of motion pairs was randomized to avoid order effects. After each trial, participants rated each motion on three subscales of the Godspeed Questionnaire Series (Anthropomorphism, Animacy, Perceived Intelligence) and four task-specific items (confidence, hesitation, rushedness, intention-to-be-captured) using a 5-point Likert scale. Written informed consent was obtained from all participants prior to the experiment.

\noindent\textbf{Results:}
Paired $t$-tests were conducted to compare the baseline motion with each phase-modified motion. The following significant differences were observed: (i) \emph{faster approach} (Phase~1) increased perceived confidence ($p<.1$); (ii) a \emph{longer stop before lifting} (Phase~2) increased perceived intelligence ($p<.05$); (iii) \emph{faster placement} (Phase~4) tended to be judged as rushed ($p<.1$); and (iv) \emph{faster withdrawal} (Phase~6) increased anthropomorphism ($p<.1$) and animacy ($p<.05$).

\subsection{Exp.~2-2: Video Condition}
Exp.~2-2 examined how motion timing is perceived when robot motions are presented via video rather than physical embodiment. The same ten participants from Exp.~2-1 completed this condition, viewing recorded videos of the seven motion patterns and responding to the same questionnaire after each trial.

\noindent\textbf{Results:}
Paired $t$-tests were conducted to compare the baseline motion with each phase-modified motion. Compared to the robot condition, effects on the Godspeed subscales (Anthropomorphism, Animacy, Perceived Intelligence) were attenuated, and no subscale-level differences reached significance. However, phase-specific effects were still observed at the task-specific items: (i) \emph{faster lifting} (Phase~3) increased perceived confidence ($p<.1$); (ii) \emph{faster placement} (Phase~4) tended to be judged as rushed ($p<.1$); and (iii) \emph{faster withdrawal} (Phase~6) increased intention-to-be-captured ($p<.05$).

\subsection{Discussion of Experiment 2}
Comparing the two conditions shows that the robot's physical presence amplifies impression changes. Significant differences were found on the Godspeed subscales mainly in the robot presentation condition, whereas video presentation produced weaker effects. This result is consistent with prior findings that embodied interaction enhances the salience of robot motion cues~\cite{bainbridge2011,mara2021,tsoi2024}.

Two implications can be drawn from these results:
\begin{itemize}
  \item \textbf{Phase-specific timing}—especially in the withdrawal phase—appears to play an important role in impression formation.
  \item \textbf{Presentation modality matters}—robot motion cues are interpreted more strongly and consistently when observed in physical form than via video.
\end{itemize}

These findings suggest that evaluations of expressive robot motion should account for embodiment effects and that timing parameters in the later motion phases warrant particular attention in motion design.

\begin{figure*}[t]
    \centering
    \begin{minipage}[b]{0.325\textwidth}
        \centering
        \includegraphics[width=0.7\textwidth]{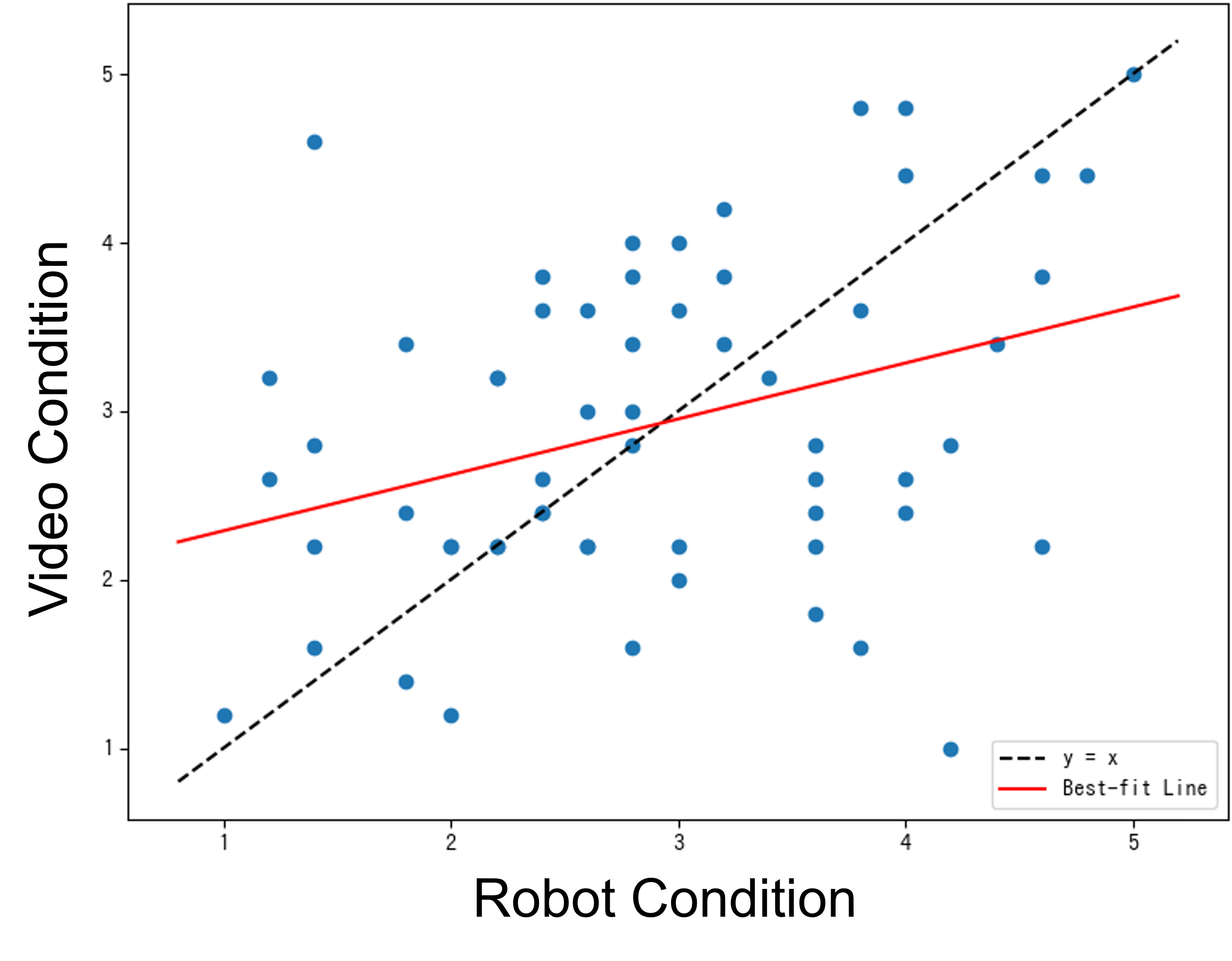}\\
        {\footnotesize (a) "Anthropomorphism" scores}
        \label{fig:sct1}
    \end{minipage}
    \begin{minipage}[b]{0.325\textwidth}
        \centering
        \includegraphics[width=0.7\textwidth]{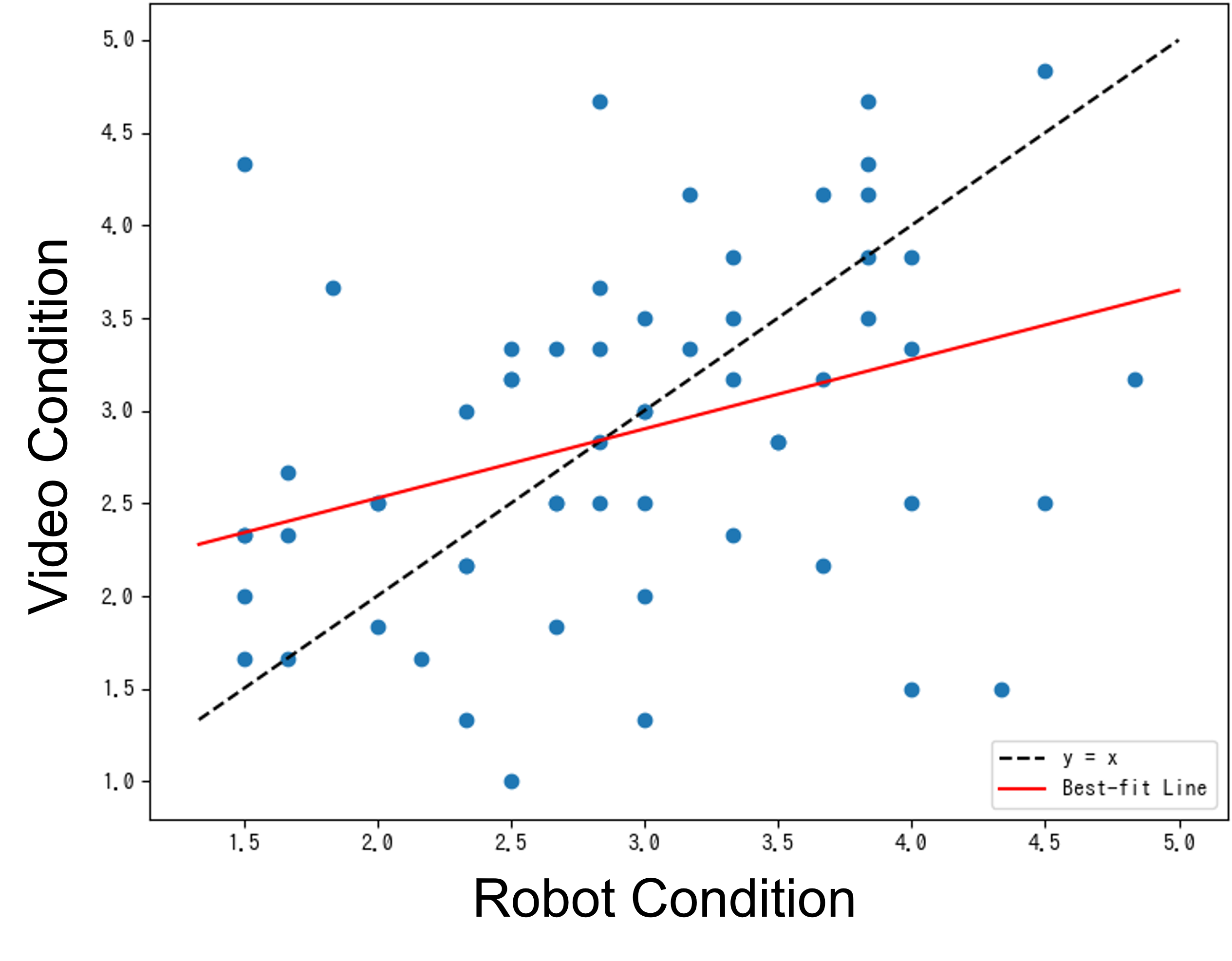}\\
        {\footnotesize (b) "Animacy" scores}
        \label{fig:sct2}
    \end{minipage}
    \begin{minipage}[b]{0.325\textwidth}
        \centering
        \includegraphics[width=0.7\textwidth]{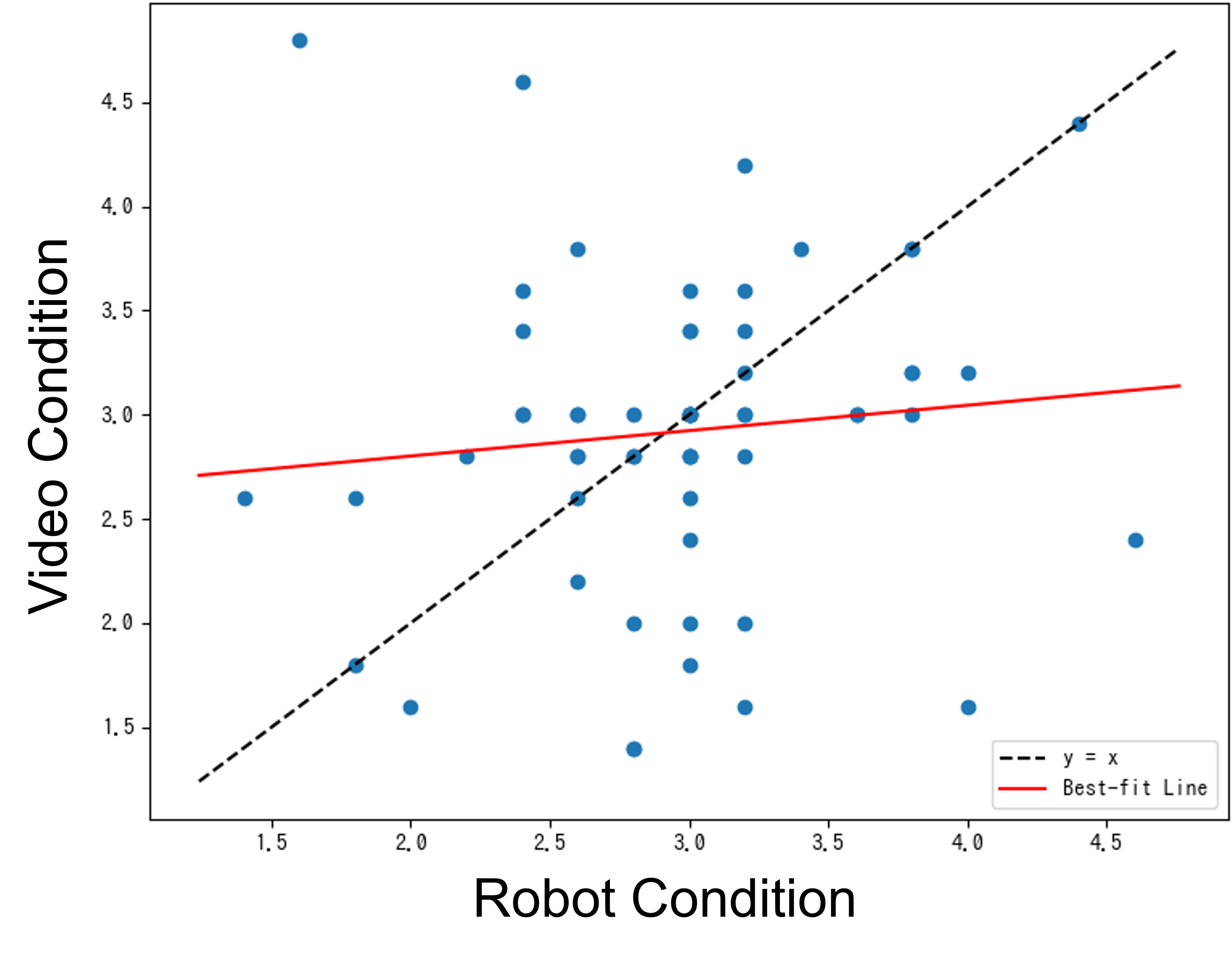}\\
        {\footnotesize (c) "Perceived Intelligence" scores}
        \label{fig:sct3}
    \end{minipage}
    \caption{Distribution of GQS Scores in Robot and Video Presentation Conditions}
    \label{fig:sct}
\end{figure*}

\section{Discussion}
\subsection{Unconscious vs. Intentional Human Movements}
Experiment~1 highlighted the contrast between natural, unconscious movements (Exp.~1-1) and intentional expressions (Exp.~1-2). 
In Exp.~1-1, later phases of gameplay showed subtle but consistent differences—slower placement, shorter pre-release stops, and faster withdrawal when moving blue pieces—indicating unconscious hesitation or cautiousness under risk. However, strategic deception in \emph{Geister} likely suppressed overt cues~\cite{depaulo2003cues}.
In contrast, Exp.~1-2 revealed that intentional expression produced clearer and more phase-specific differences, consistent with findings on motion legibility and expressivity~\cite{dragan2013legibility,Zhou2017}, though with high inter-individual variability.


These results suggest a trade-off for expressive robot motion design: unconscious patterns provide subtle but stable cues, whereas intentional expressions yield stronger but less generalizable signals. A balanced design may thus combine naturally occurring cues with selectively amplified elements to achieve both naturalness and clarity.

\subsection{Physical Robot vs. Video Presentation}
Experiment~2 confirmed that presentation modality significantly affects impression formation. 
In Exp.~2-1, physical embodiment amplified impression changes, particularly in the GQS subscales (Anthropomorphism, Animacy, and Perceived Intelligence), which reflect social presence. 
In contrast, these effects were attenuated in the video condition (Exp.~2-2), in which no subscale-level differences were observed.

These results indicate that motion cues that are salient in physical observation often weaken or are lost in video form, consistent with prior findings on embodiment effects~\cite{bainbridge2011,mara2021,tsoi2024}. 
Nevertheless, even in the video condition, task-specific intentions were still conveyed—for instance, faster withdrawal (Phase~6) increased perceived “intention-to-be-captured” ($p<.05$). 
This suggests that while embodiment enhances social presence and life-likeness, specific strategic timing cues can still be effectively communicated via video, depending on the robot’s intended social role.



\subsection{Mismatch between Human Intention and Robot Impression}
Integrating the findings from Experiments~1 and~2 suggests a potential mismatch between the strategic intentions humans express through motion and the social impressions observers form when a robot executes those motions.

For instance, in Exp.~1-2, when participants were instructed to express “wanting the opponent to capture the piece” (red piece), many exhibited a longer stop in Phase~2 (before lifting). However, when the robot performed this same action in Exp.~2-1, observers rated it as significantly higher in Perceived Intelligence ($p<.05$).

This suggests that timing cues humans use to express task-specific intentions may be interpreted by observers as broader social attributes (e.g., intelligence or deliberation) when executed by a robot. Thus, while imitating human timing patterns can effectively shape social impressions, these impressions may not align with the intended communicative meaning, posing a challenge for expressive motion design.




\subsection{Implications for Robot Motion Design}
Taken together, our findings suggest three key implications for the design of expressive robot-arm motions:

\begin{itemize}
  \item \textbf{Prioritize Late-Phase Timing:} 
  Timing variations in the later motion phases, particularly the withdrawal phase, appears to play an important role in impression formation. Late-phase timing should therefore be prioritized when designing expressive motion.

  \item \textbf{Select cues:} Unconscious human motion patterns provide subtle but sometimes ambiguous cues, whereas intentional expressions yield clearer and more interpretable signals. Such cues should be appropriately selected to enhance communicative effectiveness.


  \item \textbf{Embodiment effect:} Physical embodiment enhances the perception of motion cues compared to video presentation. Evaluations of expressive robot motion should therefore include physical interaction settings rather than relying solely on video-based assessment.
\end{itemize}

\section{Conclusion}
This study analyzed human arm movements in the imperfect-information game \emph{Geister} and examined how timing-based motion cues influence the impression of robot-arm movements. Natural gameplay reflected unconscious tendencies, whereas intentionally emphasized movements produced clearer and more distinguishable cues. In addition, impressions differed between physical and video presentation, indicating that embodiment affects the perception of expressive timing cues.

These findings suggest that consciously emphasized human timing behaviors are effective design cues for expressive robot motion, while physical embodiment should be considered in evaluation. Future work will increase participant diversity, refine instructions for intentional expression, and investigate combined phase manipulations to clarify how motion cues jointly shape perceived intention.

\section*{Acknowledgment}
This research was supported by JSPS KAKENHI Grants JP23K11278 and JP21K11968.

\balance

\bibliographystyle{IEEEtran}
\bibliography{roboarm}

\end{document}